\documentclass[10pt,twocolumn,letterpaper]{article}

\usepackage{temp}
\usepackage{times}
\usepackage{epsfig}
\usepackage{graphicx}
\usepackage{amsmath}
\usepackage{amssymb}
\usepackage{array}
\usepackage{caption}
\usepackage{multirow}
\usepackage{color}
\usepackage{makecell}

\newcolumntype{L}{>{\centering\arraybackslash}m{0.8cm}}
\newcolumntype{X}{>{\centering\arraybackslash}m{1.1cm}}
\newcolumntype{I}{>{\centering\arraybackslash}m{1.3cm}}
\newcolumntype{J}{>{\centering\arraybackslash}m{1.2cm}}
\newcolumntype{K}{>{\centering\arraybackslash}m{1.5cm}}
\newcolumntype{A}{>{\centering\arraybackslash}m{0.6cm}}
\newcolumntype{S}{>{\arraybackslash}m{2cm}}

\usepackage[pagebackref=true,breaklinks=true,letterpaper=true,colorlinks,bookmarks=false]{hyperref}

\tempfinalcopy % *** Uncomment this line for the final submission

\def\tempPaperID{000}

\iftempfinal\fi
\begin{document}

%%%%%%%%% TITLE
\title{Do Less and Achieve More: Training CNNs for Action Recognition \\ Utilizing Action Images from the Web}

\author{Shugao Ma$^{1}$,  Sarah Adel Bargal$^{1}$, Jianming Zhang$^{1}$, Leonid Sigal$^{2}$, Stan Sclaroff$^{1}$\\
$^{1}$Boston University, $^{2}$Disney Research\\
{\tt\small shugaoma@bu.edu, sbargal@bu.edu, jmzhang@bu.edu, lsigal@disneyresearch.com, sclaroff@bu.edu}
}

\makeatletter
\let\@oldmaketitle\@maketitle% Store \@maketitle
\renewcommand{\@maketitle}{\@oldmaketitle% Update \@maketitle to insert...
  \includegraphics[width=1\linewidth]{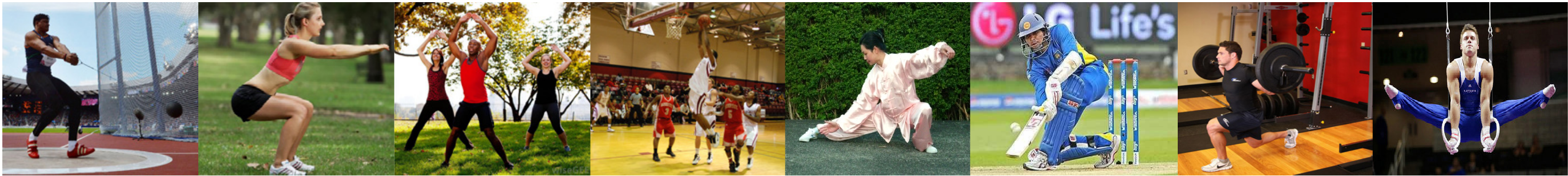}
  \captionof{figure}{Sample action images from our dataset. Action images on the Web often capture well-framed descriminative poses of the actions they represent. Left to right: {\em Hammer Throw}, {\em Body Weight Squats}, {\em Jumping Jack}, {\em Basketball}, {\em Tai Chi}, {\em Cricket Shot}, {\em Lunges}, {\em Still Rings}. Utilizing web action images in training CNNs, for all these action classes, results in more than 10\% absolute increase in recognition accuracy in videos compared to CNNs trained only on video frames (see Fig.~\ref{fig:improvActions}).}
  \label{fig:defPose}
  \bigskip\bigskip}
\makeatother

\maketitle
%\thispagestyle{empty}

%%%%%%%%% ABSTRACT
\begin{abstract}
Recently, attempts have been made to collect millions of videos to train CNN models for action recognition in videos. However, curating such large-scale video datasets requires immense human labor, and training CNNs on millions of videos demands huge computational resources. In contrast, collecting action images from the Web is much easier and training on images requires much less computation. In addition, labeled web images tend to contain discriminative action poses, which highlight discriminative portions of a video's temporal progression. We explore the question of whether we can utilize web action images to train better CNN models for action recognition in videos. We collect 23.8K manually filtered images from the Web that depict the 101 actions in the UCF101 action video dataset. We show that by utilizing web action images along with videos in training, significant performance boosts of CNN models can be achieved. % Our spatial CNN achieves 83.5\% accuracy on UCF101, which is 10.5\% absolute improvement over a spatial CNN trained only on videos. When combined with motion features, we achieve 91.1\% accuracy, which is an improvement on the state-of-the-art. 
We then investigate the scalability of the process by leveraging crawled web images (unfiltered) for UCF101 and ActivityNet. We
% provide weak supervision by auto-filtering
 replace 16.2M video frames by 393K unfiltered images and get comparable performance.

\end{abstract}

%%%%%%%%% BODY TEXT

%-------------------------------------------------------------------------------
\section{Introduction}

%\begin{figure*}[t]
%  \center
%  \includegraphics[width=1\linewidth]{sport_examples.pdf}\\
%  \caption{Sample sports images we collect from the web. Web action images, especially sports images, usually captures %discriminative poses of the actions, thus providing implicit supervision on {\em visual definition} of the actions.}
% \label{fig:sportsSample}
%\end{figure*}

Recent works \cite{KarpathyTSLSF14, simonyan2014two} show that deep convolutional neural networks (CNNs) are promising for action recognition in videos. However, CNN models typically have millions of parameters \cite{chatfield2014return, krizhevsky2012imagenet, simonyan2014very}, and usually large amounts of training data are needed to avoid overfitting. For this purpose, work is underway to construct datasets consisting of millions of videos \cite{KarpathyTSLSF14}. However, the collection, pre-processing, and annotation of such datasets can require a lot of human effort. Moreover, storing and training on such large amounts of data can consume substantial computational resources.  

In contrast, collecting and processing images from the Web is much easier. For example, one may need to look through all, or most, video frames to annotate the action, but often a single glance is enough to decide on the action in an image. Videos and web images also have complementary characteristics.
A video of 100 frames may convey a complete temporal progression of an action.
In contrast, 100 web action images may not capture the temporal progression, but do tend to provide more variations in terms of camera viewpoint, background, body part visibility, clothing, \etc. Moreover, videos often contain many redundant and uninformative frames, \eg~, standing postures, whereas action images tend to focus on discriminative portions of the action (Fig.~\ref{fig:defPose}). This property can further focus the learning, making action images inherently more valuable.

In this work, we ask the question: {\em Can web action images be leveraged to train better CNN models and to reduce the burden of curating large amounts of training videos?}  

This is not a question with an easy {\em yes} or {\em no} answer. First, web action images are usually photos, such as professional photos, commercial photos, or artistic photos, which can differ significantly from video frames. This can introduce domain shift artifacts between videos and images. Second, adding web action images in training may have different effects for different actions and for different CNN models. Furthermore, the performance improvement as a function of the Web image set size should be studied.

We start by collecting a large web action image dataset that contains 23.8K images of 101 action classes. Our dataset is more than double the size of the largest previous action image dataset \cite{yao2011human}, both in the number of images and the number of actions. And, to the best of our knowledge, this is the first action image dataset that has one-to-one correspondence in action classes with the large-scale action recognition video benchmark dataset, UCF101 \cite{ucf101}. Images of the dataset are carefully labeled and curated by human annotators; we refer to them as {\em filtered} images. Our dataset will be made publicly available for research.

For a thorough investigation, we train CNN models of different depths and analyze the effect of adding web action images to the training set of video frames for different action classes. We also train and evaluate models with varying numbers of action images to explore marginal gain as a function of the web image set size. We find that by combining web action images with video frames in training, a spatial CNN can achieve an accuracy of 83.5\% on UCF101, which is more than 10\% absolute improvement over a spatial CNN trained only on videos \cite{simonyan2014two}. When combining with motion features, we can achieve 91.1\% accuracy, which is the highest result reported to-date on UCF101. We also replace videos by images to demonstrate that our performance gains are due to images providing complementary information to that available in videos, and not solely due to additional training data.

We then further investigate how our approach can be made scalable. We crawl a dataset of web images for UCF101 from the web. These crawled images are not manually labeled; we refer to them as {\em unfiltered} images. We compare the performance of filtered and unfiltered images on UCF101. Using more unfiltered images we obtain similar performance to that obtained using fewer filtered images. We also crawl a dataset of web images for ActivityNet \cite{caba2015activitynet}; a larger scale action recognition video dataset. We obtain comparable performance when replacing half the training videos in ActivityNet (which correspond to 16.2M frames) by 393K unfiltered web images. Both crawled datasets will be made publicly available for research.

% Fine-tuning a CNN model that was pre-trained on large-scale datasets (but often for a different task) can be quite effective for initializing CNN model parameters, especially when training data for the target task is limited [reference ???].

% Zombies
% In our experimental analysis, we bring to light the downside of such a strategy: it can produce {\em zombie filters} -- CNN filters that undergo small changes during fine-tuning and make little, if any, contribution to the target task. These zombie filters reduce the number of effective parameters in the CNN model and are potentially harmful to the modeling capacity of the CNN. We investigate this issue through various experiments and analysis, which we hope will aid future work.

In summary, our \textbf{contributions} are:
\begin{itemize}
\item We study the utility of {\em filtered} web action images for video-based action recognition using CNNs. By including filtered web action images in training we improve the accuracy of spatial CNN models for action recognition by 10.5\%. 
%and, combining with motion features we achieve state-of-the-art on UCF101.
\item We study the utility of {\em unfiltered} crawled web action images, a more scalable approach, for video-based action recognition using CNNs. We obtain comparable performance when replacing half ActivityNet videos (16.2M frames) with 393K unfiltered web images.
\item We collect the largest web action image dataset to-date. This dataset is in one-to-one correspondence with the 101 actions in the UCF101 benchmark. We also collect two crawled action image datasets corresponding to the classes of UCF101 and ActivityNet. 
\end{itemize}

%-------------------------------------------------------------------------
\section{Related Work}
Action recognition is an important research topic for which a large number of methods have been proposed \cite{weinland2011survey}. Among these, due to promising performance on realistic videos including web videos and movies, bag-of-words approaches that employ expertly-designed local space-time features have been widely used. Some representative works include space-time interest points \cite{laptev08} and dense trajectories \cite{wang2013improve}. Advanced feature encoding methods, \eg~ Fisher vector encoding \cite{perronnin2010improving}, can be used to further improve the performance of such methods \cite{wang2013lear}. Besides bag-of-words approaches, other works make an effort to explicitly model the space-time structures of human actions \cite{Raptis2013, WangQT14, WangM11} by using, for example, HCRFs and MRFs. 

CNN models learn discriminative visual features at different granularities, directly from data, which may be advantageous in large-scale problems. CNN models may implicitly capture higher-level structural patterns in the features learned at the last layers of the CNN model. In addition, CNN features may also be used within structured models like HCRFs and MRFs to further improve performance. 

Some recent works propose the use of CNN models for action recognition in videos \cite{ji20133d, KarpathyTSLSF14, ng2015beyond, simonyan2014very}. Ji \etal \cite{ji20133d} use 3D convolution filters within a CNN model to learn space-time features. Karpathy \etal \cite{KarpathyTSLSF14} construct a video dataset of millions of videos for training CNNs and also evaluate different temporal fusion approaches. Simonyan and Zisserman \cite{simonyan2014very} use two separate CNN streams: one CNN is trained to model spatial patterns in individual video frames and the other CNN is trained to model the temporal patterns of actions,  based on stacks of optical flow. Ng \etal \cite{ng2015beyond} use a recurrent neural network that has long short-term memory (LSTM) cells. In all of these works, the CNN models are trained only on videos. Our findings regarding the use of web action images in training may help in further improving the performance of these works.

Web action images have been used for training non-CNN models for action recognition \cite{chen2013watching, ikizler2012web} and event recognition \cite{duan2012exploiting, wang2014annotate} in videos. Ikizler-Cinbis \etal \cite{ikizler2012web} use web action images to train linear regression classifiers for small-scale action classification tasks (5 or 8 action classes). Chen \etal \cite{chen2013watching} use static action images to generate synthetic samples for training SVM action classifiers and evaluate on a small test set of 78 videos comprising 5 action classes. In \cite{duan2012exploiting}, Duan \etal use SVMs trained on SIFT features of web action images in their video event recognition system and evaluate on datasets with 5$\sim$6 different events. Wang \etal \cite{wang2014annotate} exploit semantic groupings of Web images for video event recognition and evaluate on the same datasets as  \cite{duan2012exploiting}. Sun \etal \cite{sun2015temporal} localize actions temporally using a domain transfer from web images. In contrast, our work gives the first thorough study on combining web action images with videos for training CNN models for large-scale action recognition.

%--------------------------------------------------------------------------
\section{Web Action Image Dataset}
\label{sec:dataset}

To study the usefulness of web action images for learning better CNN models for action recognition, we collect action images that correspond with the 101 action classes in the UCF101 video dataset.

For each action class, we automatically download images from the Web  (Google, Flickr, etc.) using corresponding key phrases, \eg~ {\em pushup training} for the class {\em pushup}, and then manually remove irrelevant images or drawings and cartoons. We also include 2769 images of relevant actions from the Standford40 dataset \cite{yao2011human}.  The resulting dataset comprises 23.8K images. Because the images are automatically collected, and then filtered for irrelevant ones, the number of images per category varies. Each class has at least 100 images and most classes have 150-300 images. We will make our dataset publicly available for research.

\begin{table}
\footnotesize
\center
\caption{Comparison of our action image dataset with existing action image datasets. {\em Visibility varies?} refers to variance in the partial visibility of the human bodies.}
\begin{tabular}{S | L L L A L}
\hline \hline
Dataset & No. of actions & No. of images & Clutter? & Poses vary? & Visibility varies? \\ \hline
Gupta \cite{gupta2009observing}     & 6 & 300 & Small & Small & No \\
Ikizler \cite{ikizler2009learning}  & 5 & 1727 & Yes & Yes & Yes \\ 
VOC2012 \cite{everingham2010pascal} & 11 & 4588 & Yes & Yes & Yes \\ 
PPMI \cite{yao2010grouplet}         & 24 & 4800 & Yes & Yes & No \\ 
Standford40 \cite{yao2011human}     & 40 & 9532 & Yes & Yes & Yes \\ 
\textbf{Ours}                       & \textbf{101} & \textbf{23800} & \textbf{Yes} & \textbf{Yes} & \textbf{Yes} \\ 
\hline \hline
\end{tabular}
\label{tab:imgDatasets}
\end{table}      

Table~\ref{tab:imgDatasets} compares existing action image datasets with our new dataset. Both in the number of images and the number of actions, our dataset exceeds double the scale of existing datasets. More importantly, to the best of our knowledge, this is the first action image dataset that has one-to-one action class correspondence with a large-scale action recognition benchmark video dataset. We believe that our dataset will enable further study of the relationship between action recognition in videos and in still images. 
%\begin{figure}[t]
%  \center
%  \includegraphics[width=0.9\linewidth]{boxplot_dotted.png}\\
%  \caption{Box plot of the distribution of image numbers in action classes of different types}
% \label{fig:imgNum}
%\end{figure}

%Because the images are automatically collected, and then filtered for irrelevant ones, the number of images per category varies. 
% Each class has at least 100 images and most classes have 150-300 images. 
%
%Fig.~\ref{fig:imgNum} summarizes the distribution of the number of images per class, in our dataset, among these action category types.  
%It is important to note that due to the way we collected these images, this distribution is also reflects the number of images we may collect from the Web for these actions.
%
%More detailed statistics are provided in the supplementary material. 
% More detailed statistics about the number of images per class and 
%the distribution of image resolutions per class 
% image quality are provided in the supplementary material.

\begin{figure}[t]
  \center
  \includegraphics[width=1\linewidth]{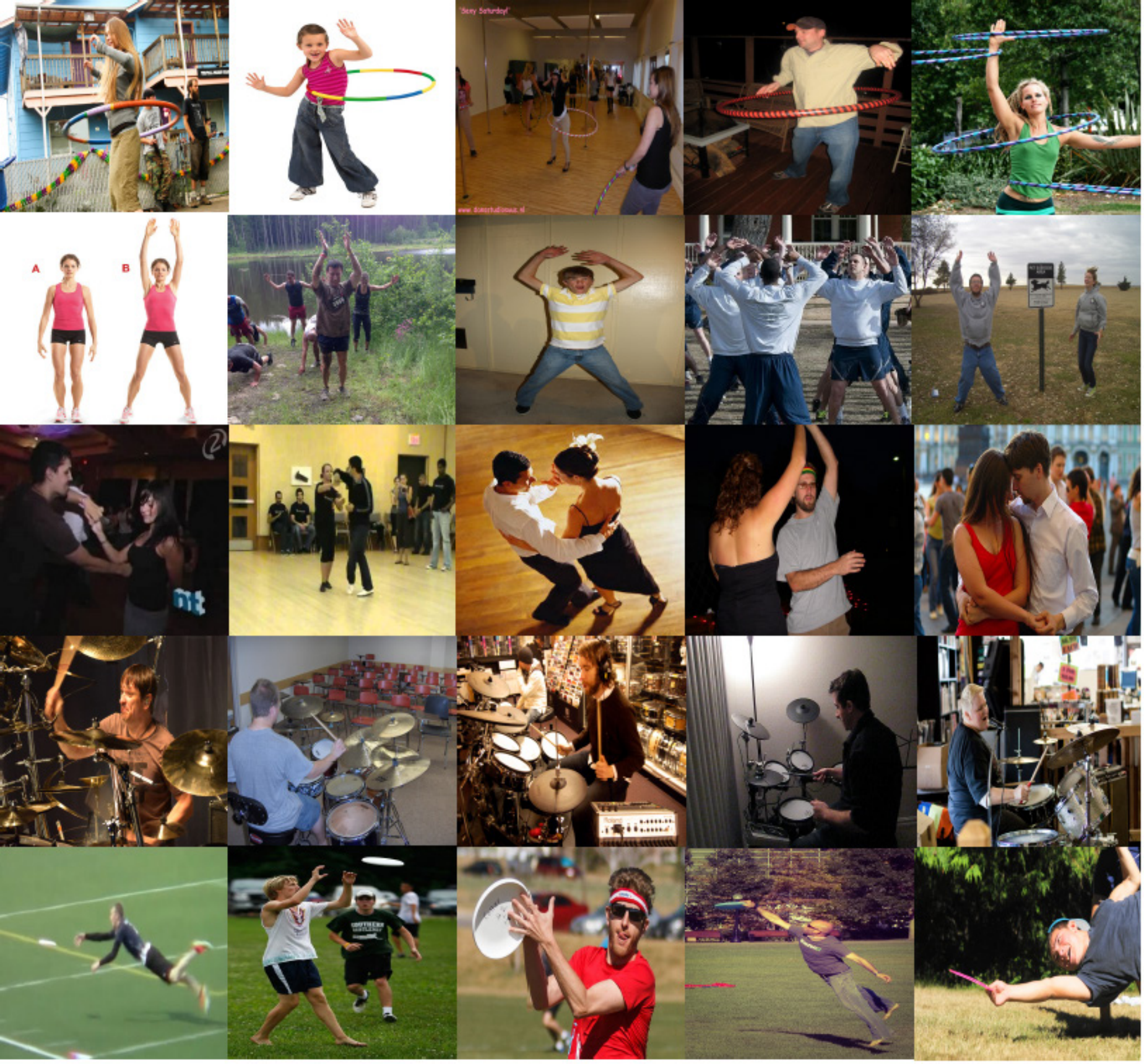}\\
  \caption{Sample images from our action image dataset. Each row shows images of one action. Top to bottom: {\em Hula Hoop, Jumping Jack, Salsa Spin, Drumming, Frisbee Catch}. Variations in background, camera viewpoint and body part visibility are common in web images of the same action.}
 \label{fig:imgSample}
\end{figure}

UCF101 action classes are divided into five types: {\em Human-Object Interaction},  {\em Body-Motion Only}, {\em Human-Human Interaction},  {\em Playing Musical Instruments}, and {\em Sports} \cite{soomro2012ucf101}. Fig.~\ref{fig:imgSample} shows sample images in our dataset for five action classes, one in each of the five action types. 

These action images collected from the Web are originally produced in a variety of settings, such as amateur \vs professional photos, artistic \vs educational \vs commercial photos, etc. For images collected in each action category, wide variation can exist in viewpoint, lighting, human pose, body part visibility, and background clutter. For example, commercial photos may have clear backgrounds while backgrounds of amateur photos may contain much more clutter. Such variance also differs for different types of actions. For example, for {\em Sports}, there is significant variance in body pose among images that capture different phases of the actions, whereas body pose variance is minimal in images of {\em Playing Musical Instruments}. %Furthermore, 

Many of the collected action images significantly differ from video frames in camera viewpoint, lighting, human pose, and background. One interesting thing to notice is that action images often capture {\em defining poses} of an action that are highly discriminative, \eg~ standing with both hands over head and legs spread in {\em jumping jack} (Fig.~\ref{fig:imgSample}, row 2). In contrast, videos may have many frames containing poses that are common to many actions, \eg~ in {\em jumping jack} the upright standing pose with hands down. Also, $n$ images will have more unique content than $n$ video frames, for example more clothing variation. Clearly there exists a compromise between temporal information available in videos and discriminative poses and variety of unique content in images.

%-------------------------------------------------------------------------
\section{Training CNNs with Web Action Images}
\label{sec:training}

\begin{figure*}[t]
  \center
  \includegraphics[width=1\linewidth]{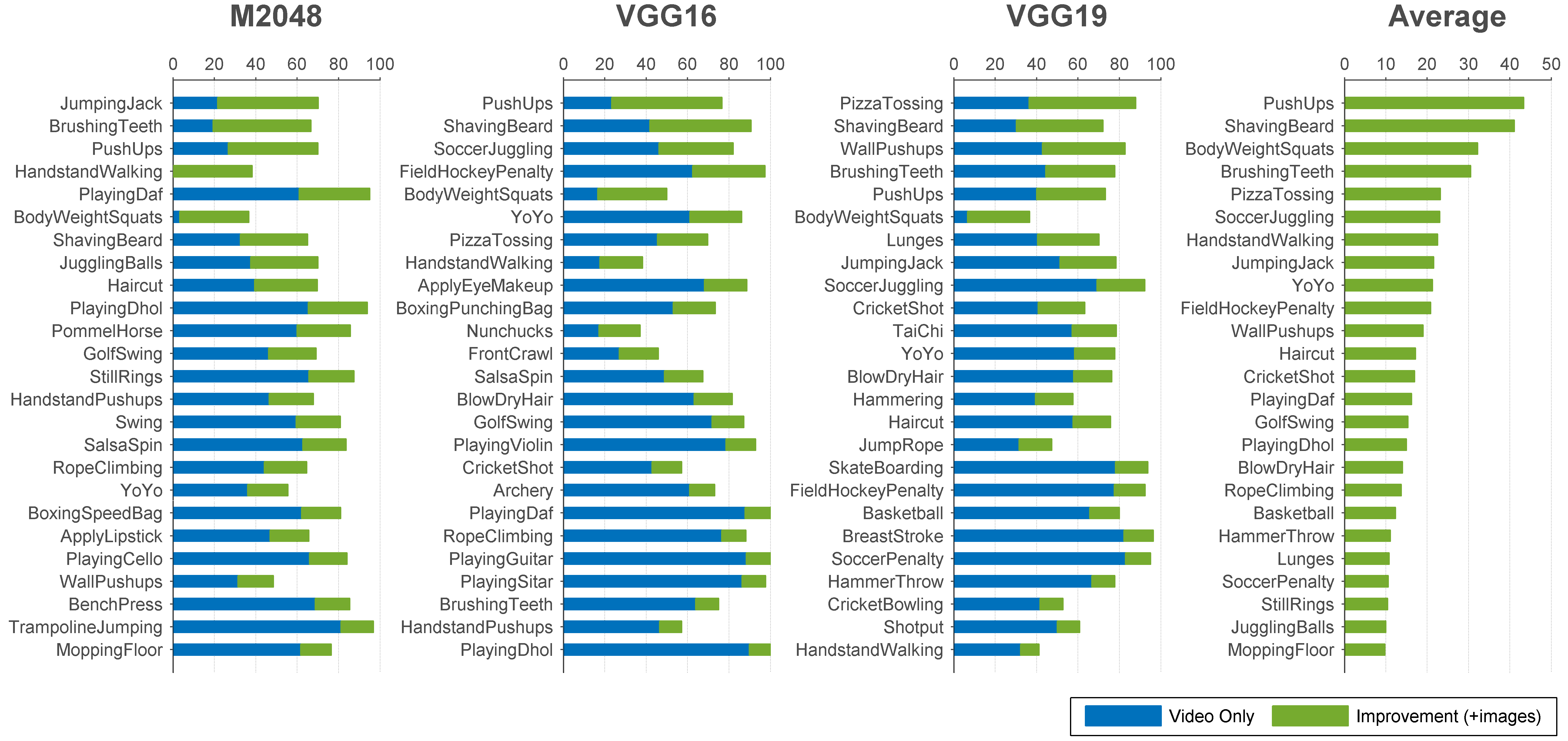}\\
  \caption{The 25 action classes with the largest accuracy improvement in the three CNN architectures as well as on average over the three architectures. The blue bars show the accuracy of CNN models trained only on videos. The green bars show the absolute increase in accuracy of CNN models  trained using both web action images and training videos.} 
 \label{fig:improvActions}
\end{figure*} 

Spatial CNNs trained on single video frames for action recognition are explored in \cite{simonyan2014two}. Karpathy \etal \cite{KarpathyTSLSF14} observe that spatio-temporal networks show similar performance compared to spatial models. A spatial CNN effectively classifies actions in individual video frames, and action classification for a video is accomplished via fusion of the spatial CNN's outputs over multiple frames, \eg~via voting or SVM. Because the spatial CNN is trained on single video frames, its parameters can be learned by fine-tuning of a CNN that was trained for a different task, \eg~, using a CNN that is pre-trained on ImageNet \cite{deng2009imagenet}. 
The fine-tuning approach is especially beneficial in training a CNN model for action classification in videos, since we often only have limited training samples; given the large number of parameters in a CNN, initializing the parameters to random values leads to overfitting and inferior performance as shown in \cite{simonyan2014two}. In this work, we study improving the spatial CNN for action recognition using web action images as training data in fine-tuning. This is then combined with motion features via state-of-the-art techniques. 

In our experiments and analysis, we explore the following key questions:
\begin{itemize}
\item Is it beneficial to train CNNs with web action images in addition to video frames and, if so, which action classes benefit most?
\item How do different CNN architectures, in particular ones with different depths, perform when web action images are used as additional training data?
\item How do the performance gains change when more web action images are used in training the CNN?
\item Are performance gains solely due to additional training data or also due to a single image being more informative than a randomly sampled video frame?
\item Can we make the procedure of leveraging web images scalable by using crawled (unfiltered) web images rather than manually filtered ones?
%and, can we auto-filter the crawled images to reduce the number of training samples yet maintain performance?
\end{itemize}

We experiment on three CNN architectures: M2048 \cite{chatfield2014return}, VGG16, and VGG19 \cite{simonyan2014very}. To avoid cluttering the discussion, implementation details are provided later in Sec.~\ref{sec:experiment}.

\vspace{0.1in}
\noindent \textbf{Is adding web images beneficial?} 
Significant performance gains are achieved when we train spatial CNNs using our web action image dataset as auxiliary training data (see Table~\ref{tab:modelComp}). For example, with the VGG19 CNN architecture, 5.7\% absolute improvement in mean accuracy is achieved.

Most encouragingly, such improvements are easy to implement, without the need to introduce additional complexity to the CNN architecture and/or requiring significantly longer training time. 

We further analyze which classes improve the most. Fig.~\ref{fig:improvActions} shows the 25 action classes for which the largest improvement in accuracy is achieved with the three different CNN architectures on UCF101 split1. The 25 action classes of top average accuracy improvement over all three tested architectures are also shown (rightmost column), all of which have no less than 10\% absolute increase in accuracy and 10 classes have more than 20\% absolute improvement. Some action classes are consistently improved irrespective of the CNN architecture used, such as {\em push ups}, {\em YoYo}, {\em handstand walking}, {\em brushing teeth}, {\em jumping jack}, etc. This suggests that utilizing web action images in CNN training is widely applicable. %The average improvement for each of the 101 classes is included in the supplemental material; 80 classes showed some improvement, 49 of which showed an improvement of at least 5\%.

While classification accuracy improvements in actions that are relatively {\em stationary} such as {\em Playing Daf} and {\em Brushing Teeth} are somewhat expected, it is interesting to see that improvements for actions of fast body motion such as {\em Jumping Jack} and {\em Body Weight Squats} are also significant. 

\begin{table}
\centering
\caption{Accuracy on UCF101 split1 using three different CNN architectures.}
\begin{tabular}{l | X I J K}
\hline \hline
Model  & \# layers & \# param. {\scriptsize (in Millions)} & Accuracy {\scriptsize video only} & Accuracy {\scriptsize video + images}\\ \hline 
M2048  &   7       &  91                                   & 66.1\%    &  \textbf{75.2\%} \\ 
VGG16  &  16       &  138                                  & 77.8\%    &  \textbf{83.5\%}       \\ 
VGG19  &  19       &  144                                  &  78.8\%   &  \textbf{83.5\%}   \\ 
\hline \hline
\end{tabular}
\label{tab:modelComp}
\end{table}

\vspace{0.1in}
\noindent \textbf{Are images benefitial irrespective of CNN depth?}
While there are numerous ways that CNN architectures may differ from each other, here we focus on one of the most important factors. We evaluate the performance changes for CNNs of different depths when web action images are used in addition to video frames in training. We train spatial CNNs of three depths: 7 layers (M2048), 16 layers (VGG16) and 19 layers (VGG19). These are the prototypical choices of CNN depths in recent works \cite{chatfield2014return, krizhevsky2012imagenet, long2014fully, simonyan2014two, simonyan2014very}.

Table~\ref{tab:modelComp} shows the mean accuracy of the three CNN models trained {\em with} and {\em without} web action images on UCF101 split1. Using web action images in training leads to a consistent 5\% $\sim$ 9\% absolute improvement for all three architectures of different depths. This shows the usefulness of web action images and suggests a wide applicability of this approach. Furthermore, our results in action recognition confirm \cite{simonyan2014very}'s observation that deeper CNNs of 16-19 layers significantly outperform the shallower 7-layer architecture. However, the margin of performance gain diminishes when we increase the depth from 16 to 19. 

\vspace{0.1in}
\noindent \textbf{Does adding more web images improve accuracy?} We further explore how, for the same CNN architecture, the number of web action images used as additional training data can influence the classification accuracy of the resulting CNN model. We sample $1/10$, $1/5$, $1/3$ and $2/3$ of the images of each action in our dataset, and for each sampled set we train the spatial CNN by fine-tuning VGG16 using both the training videos and sampled action images. For each sample size, we repeat the experiment three times, each with a different randomly sampled set of web action images. The evaluation is performed on UCF101 split1. 

\begin{figure}[t]
  \center
\includegraphics[width=1\linewidth, height=40mm]{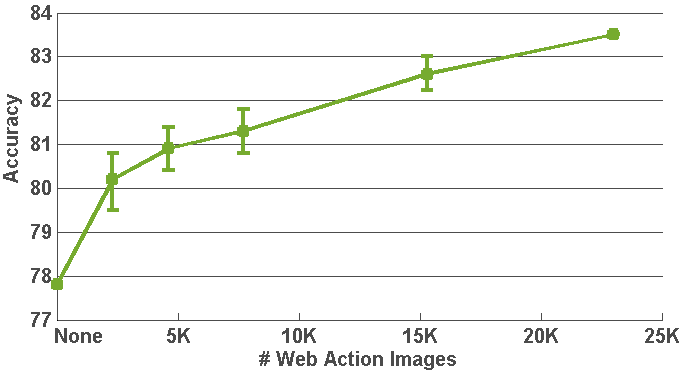}\\  
  \caption{Performance of the spatial CNNs (VGG16) trained on UCF101 split1 using different numbers of web action images as additional training data.}
 \label{fig:perfNum}
\end{figure}

Fig.~\ref{fig:perfNum} summarizes the results of this experiment. The increase in classification accuracy is most significant at the beginning of the curve, \ie~when a few thousand web action images are used in training. This increase continues as more web action images are used, even though the increase becomes slower. Firstly, this indicates that using web action images in training can make a significant difference in performance by providing additional supervision to that provided by video frames. Secondly, it indicates that it is good practice to collect a moderate number of web action images for each action as a cost-effective way to boost model performance (\eg~, 100 $\sim$ 300 images per action for a dataset of the same scale as UCF101). 
%adding more web action images in training may indeed help to improve accuracy, but 

%Thus, for a dataset of this size, we believe it is good practice to collect a moderate number of web action images for each action (\eg 100 $\sim$ 300) as a very cost-effective way to boost model performance.   

\vspace{0.1in}
\noindent \textbf{Do web images complement video frames?} Although augmenting with images is more efficient than augmenting with videos, we further investigate whether the achieved performance gains are solely due to additional training data or whether a web image provides more information to the learning algorithm than a video frame. This is done by replacing video frames by web images, keeping the total number of training samples constant. For each sample size, we repeat the experiment three times, each with a different randomly sampled set of web action images. The evaluation is performed on UCF101 split1 and a VGG16 model. 

\begin{figure}[t]
  \center
\includegraphics[width=1\linewidth]{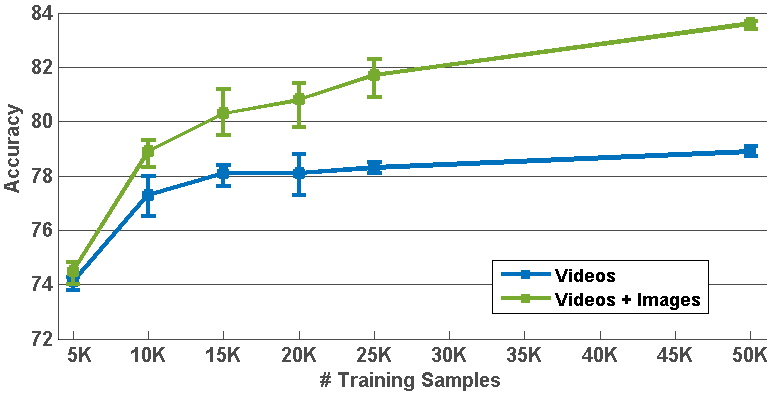}\\  
  \caption{Performance of the spatial CNNs (VGG16) trained on UCF101 split1 using video frames only and replacing 50\% of the video frames by web images.}
 \label{fig:compExp}
\end{figure}

Fig.~\ref{fig:compExp} summarizes the results of this experiment. A consistent improvement in performance is achieved when half the video frames are replaced by web images. The number of training samples (images and video frames) required to obtain the maximum accuracy presented in Fig.~\ref{fig:perfNum} is much less (50K \vs 230K). This suggests that images are augmenting the information learnt by the classifier. We posit that discriminative poses in action images may provide implicit supervision, in training, to help learn better discriminative models for classification.  

% To explore this further, we use t-SNE \cite{van2008visualizing} to project the conv53 features of images and video frames in 2D for visualization. Fig. \ref{fig:distnExp} shows how features of images and video frames are distributed for the action {\em Balance Beam}. We see the same pattern for all actions of UCF101. Distribution plots for all actions are provided in the supplemental material.

% \begin{figure}[t]
%   \center
% \includegraphics[width=1\linewidth]{BalanceBeam_ScatterPlot.png}\\  
%   \caption{The t-SNE 2D embedding of conv53 features of images and UCF101 frames for the action {\em Balance Beam}.}
%  \label{fig:distnExp}
% \end{figure}

\vspace{0.1in}
\noindent \textbf{Can this be made scalable?} While we have demonstrated the ability to collect a filtered dataset for our desired classes, this is not scalable. Given a different dataset having the same order of magnitude as UCF101 we would have to manually label a dataset for its classes. Given an even larger dataset with more classes and more samples per class, this becomes very cumbersome although still better than collecting videos. 
We now investigate the possibility of using crawled (unfiltered) web images for the same purpose. We assume that more images will be required if they are unfiltered, and so we crawl 207K unfiltered images from the Web corresponding to the classes of UCF101.  

%We then investigate whether we can auto-filter the crawled images to reduce the training complexity. Auto-filtering is performed by passing the images through the model trained on video frames only and thresholding on the confidence for the correct class. The evaluation is performed on UCF101 split1 and a VGG16 model.  

Table~\ref{tab:filtUnfilt} summarizes the results of this experiment. The performance of using unfiltered images approaches that of manually filtered images, but the number of web images utilized is much larger. We further investigate whether {\em all} the crawled unfiltered images are required to obtain such performance. We do this by randomly selecting one quarter (65.5K) of the 207K unfiltered web images. We select 3 random samples and report the average result in Table~\ref{tab:filtUnfilt}. Three quarters of the images only contribute with an additional accuracy of 1\%; this is consistent with Fig.~\ref{fig:perfNum} observations.

%The performance of using auto-filtered web images approaches that of manually filtered images with approximately 3 times as many images only.

Having demonstrated the feasibility of using crawled web images, we now apply this to a larger-scale dataset: ActivityNet \cite{caba2015activitynet}. ActivityNet contains more classes (203) and more samples per class than UCF101. ActivityNet classes are more diverse; they belong to the categories: {\em Personal Care, Eating and Drinking, Household, Caring and Helping, Working, Socializing and Leisure, and Sports and Exercises.} ``ActivityNet provides samples from 203 activity classes with an average of 137 untrimmed videos per class and 1.41 activity instances per video, for a total of 849 video hours." \cite{caba2015activitynet} Mostly, videos have a duration between 5 and 10 minutes and have a 30 FPS frame rate. About 50\% of the videos are in HD resolution. We crawl 393K unfiltered images from the Web corresponding to the classes of ActivityNet. Results on ActivityNet are reported in Section 5.

\begin{table}[t]
\centering
\caption{Accuracy on UCF101 split1 using spatial CNN (VGG16) of manually filtered and unfiltered web images.\\ * {\footnotesize means average of three random sample sets.} }
\begin{tabular}{l | c c}
\hline \hline
Image Type                 & \# Images     & Accuracy (\%) \\ \hline 
Manually filtered          & 23.8K         &  83.5  \\ 
Unfiltered (all)                & 207K          &  83.1  \\ 
% Auto-filtered $(T=0.2)$    & 65.5K         &  83.0  \\ 
Unfiltered (rand select)          & 65.5K         &  { }{ }82.1*  \\ 
% Auto-filtered $(T=0.4)$    & 56.7K         &  81.9  \\ 
% Auto-filtered $(T=0.6)$    & 49.5K         &  81.7  \\ 
\hline \hline
\end{tabular}
\label{tab:filtUnfilt}
\end{table}

%-------------------------------------------------------------------------
\section{Experiments} 
\label{sec:experiment}

\noindent Using insights from the experiments performed on UCF101 split1 in Section 4, we now perform experiments following the standard evaluation protocol \cite{THUMOS13} and report the average accuracy over the three provided splits. 

We also perform experiments on ActivityNet. Following \cite{caba2015activitynet}, we evaluate classification performance on both trimmed and untrimmed videos. Trimmed videos contain exactly one activity. Untrimmed videos contain one or more activities. We use the mAP (mean average precision) in evaluating performance. Results reported on ActivityNet are produced using the validation data, as the authors are reserving the test data for a potential future challenge.

\subsection{Implementation}
\subsubsection{Experimental Setup for UCF101}
\noindent \textbf{Fine-tuning:} We use the Caffe \cite{jia2014caffe} software for fine-tuning CNNs. We use models VGG16, VGG19 \cite{simonyan2014very}, and M2048 \cite{chatfield2014return} that are pre-trained on ImageNet by the corresponding authors.
We only test M2048 on the first split for analysis, as it is shown to be significantly inferior to the other two architectures (Table~\ref{tab:modelComp}). Due to hardware limitations, we use a small batch size: 20 for M2048 and 8 for VGG16 and VGG19.
Accordingly, we use a smaller learning rate than those used in  \cite{chatfield2014return, simonyan2014very}. For M2048, the initial learning rate $10^{-3}$ is changed to $10^{-4}$ after 40K iterations; training stops at 80K iterations. For both VGG16 and VGG19, the initial learning rate $10^{-4}$ is changed to $10^{-5}$ after 40K iterations, and is further lowered to $2\times10^{-6}$ after 80K iterations. Training stops at 100K iterations. 
Momentum and weight decay coefficients are always set to 0.9 and $5\times10^{-4}$. In each model, all layers are fine-tuned except the last fully connected layer which has to be changed to produce output of 101 dimensions with initial parameter values sampled from a zero-mean Gaussian distribution with $\sigma=0.01$.

We resize video frames to 256$\times$256, and random crops to 224$\times$224 with random horizontal flipping for training. For web action images, since their aspect ratios vary significantly, we first resize the short dimension to 256 while keeping the aspect ratio, and subsequently crop six $256\times256$ patches along the longer dimension in equal spacing. Random cropping of 224$\times$224 with random horizontal flipping is further applied to these image patches in training. Equal numbers of web images and video frames are sampled in each training batch.

\vspace{0.1in}
\noindent \textbf{Video Classification:} 
A video is classified by fusing over the CNN outputs for the individual video frames.  For a test video, we select 20 frames of equal temporal spacing. From each of the frames, 10 samples are generated following \cite{krizhevsky2012imagenet}: four corners and the center (each is 224$\times$224) are first cropped from the 256$\times$256 frame, making 5 samples;  horizontal flipping of these samples makes another 5. Their classification scores are averaged to produce the frame's scores. We classify each frame to the class of the highest score, and the class of the video is then determined by voting of the frames' classes.  

We also test SVM fusion, concatenating the CNN outputs for the 20 frames (averaged over the 10 cropped and flipped samples) from the second fully-connected layer (fc7), \ie the 15th layer in VGG16 and 18th layer in VGG19. This produces a vector of 81,920 ($4096 \times 20$) dimensions, which is then L2 normalized. One-vs-rest linear SVMs are then trained on these features for video classification. The SVM parameter $C=1$ in all experiments.

\vspace{0.1in}
\noindent \textbf{Combining with Motion Features:} The output of spatial CNNs can be combined with motion features to achieve significantly better performance, as shown in \cite{simonyan2014two}. 
%Applying our spatial CNNs in the framework of \cite{simonyan2014two} is straightforward. 
We present an alternative by combining the output of the spatial CNNs with the conventional expert-designed features, namely the improved dense trajectories with Fisher encoding (IDT-FV) \cite{wang2013lear}. We follow the same settings in \cite{wang2013lear} to compute the IDT-FV for each video except that we do not use a space-time pyramid. The IDT-FV of each video is then combined with the concatenated fc7 outputs of 20 frames to form the final feature vector for a video. One-vs-rest linear SVMs are then trained on these features for video classification. The SVM parameter $C=1$.

\begin{table} [t]
\centering
\caption{Mean accuracy of spatial CNNs (averaged over three splits) on UCF101. }
\begin{tabular}{c c}
\hline \hline
Model                                     & Accuracy (\%) \\  \hline 
slow fusion CNN \cite{KarpathyTSLSF14}    &   65.4 \\
spatial CNN \cite{simonyan2014two}        &   73.0 \\ [1ex]
VGG16, voting                             &   77.9 \\ 
VGG16 + Images, voting                    &   82.5 \\ 
VGG16 + Images, SVM fusion on fc7         &   \textbf{83.5}  \\[1ex]
VGG19, voting                             &   77.8  \\ 
VGG19 + Images, voting                    &   83.3  \\ 
VGG19 + Images, SVM fusion on fc7         &   83.4  \\ 
\hline \hline
\end{tabular}
\label{tab:resultsUCF101}
\end{table} 

%\begin{table}
%\center
%\begin{tabular}{|c|c|}
%\hline
%Model                              & HMDB \\  \hline
%VGG16, average fusion              &  \\ \hline
%VGG16 + Images, average fusion     & \\ \hline
%VGG16 + Images, SVM fusion on FC7  &  \\ \hline
%\end{tabular}
%\caption{Mean accuracy (averaged over 3 splits) on HMDB using VGG16 architecture.}
%\label{tab:resultsHMDB}
%\end{table}

\begin{table} [t]
\centering
\caption{Mean accuracy (averaged over three splits) when combining spatial CNNs with motion features for UCF101.}
\label{tab:resultsMotion}
\begin{tabular}{c c}
\hline \hline
Model                                    & Accuracy (\%) \\  \hline 
IDT-FV     \cite{wang2013lear}           & 85.9 \\ 
Two-stream CNN \cite{simonyan2014two}   & 88.0 \\ 
RCNN using LSTM \cite{ng2015beyond}      & 88.6 \\
VGG16 + Images + IDT-FV                  & \textbf{91.1} \\ 
VGG19 + Images + IDT-FV                  & 90.8 \\ 
\hline \hline
\end{tabular}
\vspace{-6pt}
\end{table}

\subsubsection{Experimental Setup for ActivityNet}
We use the Caffe \cite{jia2014caffe} software for fine-tuning CNNs. We use a VGG19 model \cite{simonyan2014very} that is pre-trained on ImageNet by the authors. Due to hardware limitations, we use a small batch size of 8. Accordingly, we use a smaller learning rate than \cite{simonyan2014very}. The initial learning rate $10^{-4}$ is changed to $10^{-5}$ after 80K iterations. Training stops at 160K iterations. Momentum and weight decay coefficients are set to 0.9 and $5\times10^{-4}$. All layers are fine-tuned except the last fully connected layer which has to be changed to produce output of 203 dimensions with initial parameter values sampled from a zero-mean Gaussian distribution with $\sigma=0.01$.

\begin{table*} [th]
\centering
\caption{Although ActivityNet is large-scale, using unfiltered web images still helps in both trimmed and untrimmed classification. * {\footnotesize means average of three random sample sets.} }
\label{tab:actNet1}
\begin{tabular}{c c c c}
\hline \hline
Model                                   & \# Images & \makecell{Untrimmed Classification \\ mAP (\%)} & \makecell{Trimmed Classification \\ mAP (\%)} \\  \hline 
fc8     \cite{caba2015activitynet}      & none      & 25.3           & 38.1 \\ 
DF \cite{caba2015activitynet}           & none      & 28.9           & 43.7 \\
Ours (video frames only)				& none      & 52.3			 & 47.7 \\
Ours (unfiltered: all)                       & 393K      & 53.8           & 49.5 \\
% Ours (auto-filtered T=0.2)              & 103K      & 53.6       	 & 49.1 \\
Ours (unfiltered: rand select)                      & 103K      & { }{ }53.3*       	 & { }{ }49.3* \\
% Ours (1/2 videos + unfiltered)			& 393K      & \textcolor{red}{[running]}		 & \textcolor{red}{[running]} \\
%Ours (auto-filtered T=0.4)				& 67.8K     & 53.3			 & \textcolor{red}{[running]} \\
%Ours (auto-filtered T=0.6)				& 44.7K     & 53.2			 & \textcolor{red}{[running]} \\
\hline \hline
\end{tabular}
\vspace{-6pt}
\end{table*}

Resizing and cropping of images and frames are performed in the same way as previously described for UCF101. Samples in each training batch are randomly selected from web action images and video frames with equal probability. 

\subsection{Results}
\subsubsection{Experimental Results for UCF101}

Here we report the performance of our spatial CNNs averaged over three splits of UCF101 (Table~\ref{tab:resultsUCF101}), as well as the performance of our models when motion features are also used (Table~\ref{tab:resultsMotion}). 

As seen in Table~\ref{tab:resultsUCF101}, all our spatial CNNs trained using both videos and images improved $\sim$10\% (absolute) in accuracy over the spatial CNN of \cite{simonyan2014two}, which is a 7-layer model. We believe this improvement is due to two main factors: using a deeper model and using web action images in training. Comparing the performance of the spatial CNN of \cite{simonyan2014two} to the deeper models trained only on videos (rows 3 and 6 in Table~\ref{tab:resultsUCF101}), we find that the improvements solely due to differences of CNN architectures are 4.9\% and 4.8\% for VGG16 and VGG19 respectively. When web action images are used in addition to videos in training (rows 4 and 7 in Table~\ref{tab:resultsUCF101}), these improvements are doubled: 9.5\% and 10.3\% respectively.

Results reported in Table~\ref{tab:resultsUCF101} show that, in the models we tested, the simple approach of using web action images in training contributes at least equally with introducing significant complexities to the CNN model, \ie~, adding at least 9 more layers. It is also interesting to note that, without using optical flow data, our spatial CNNs already approach performance attained using state-of-the-art expert designed features that use optical flow, \ie IDT-FV \cite{wang2013lear} in Table~\ref{tab:resultsMotion}. Performance gains obtained by our approach are especially encouraging compared to deepening the model or incorporating motion features, as leveraging web images during training will not add any additional computational or memory burden during test time.

The slow fusion CNN \cite{KarpathyTSLSF14} is not a spatial CNN as it is trained on multiple video frames instead of single video frames. We list it here as it presents a different approach; collecting millions of web videos for training. However, despite the fact that 1M web videos are used as pre-training data, its performance is far lower than our models. 

We further test the features learned by our spatial CNNs when combined with motion features, \ie Fisher encoding on improved dense trajectories. Table~\ref{tab:resultsMotion} compares our results with state-of-the-art methods that also use motion features. Our method (VGG16 + Images + IDT-FV) outperforms all, improving by 2.5\% over \cite{ng2015beyond} that trains recurrent CNNs with long short-term memory cells; by 3.1\% over \cite{simonyan2014two}, which combines two separate CNNs trained on video frames and optical flow respectively; and by 5.2\% over \cite{wang2013lear} that uses Fisher encoding on improved dense trajectories.

\subsubsection{Experimental Results for ActivityNet}
Here we report the performance of our spatial CNNs on ActivityNet for the task of action classification in trimmed and untrimmed videos with and without auxiliary web images (Table~\ref{tab:actNet1}). We then further investigate the use of web images as a substitute for many training videos (Table~\ref{tab:actNet2}).

\begin{table} [t]
\centering
\caption{Comparable performance is achieved when half the training videos of ActivityNet are replaced by 393K images (row 4 \vs row 1). * {\footnotesize means average of three random sample sets.} }
\label{tab:actNet2}
\begin{tabular}{c c c c}
\hline \hline
Experiment        & \# Frames & \# Images &  mAP (\%) \\  \hline 
All vids          & 32.3M     & none      & 47.7 \\
1/2 vids          & 16.2M     & none      & { }{ }40.9* \\ 
1/4 vids          & 8.1M      & none	  & { }{ }33.4* \\
1/2 vids + imgs	  & 16.2M     & 393K      & { }{ }46.3* \\
1/4 vids + imgs	  & 8.1M      & 393K      & { }{ }41.7* \\
\hline \hline
\end{tabular}
\end{table}

In Table~\ref{tab:actNet1} we observe that utilizing web images still helps $\sim$1.5\% even with a very large scale dataset like ActivityNet. Using a random sample of approximately one quarter of the crawled web images gives nearly the same results, suggesting that performance gains diminish as the number of web action images greatly increase. This result is consistent with results on UCF101 (Figure~\ref{fig:perfNum}).

In Table~\ref{tab:actNet2} we observe that comparable performance is achieved when half the training videos, are replaced by web images (rows 1 and 4 in Table~\ref{tab:actNet2}). A similar pattern is observed when repeating the experiment at a smaller scale. This suggests that using a relatively small number of web images can help us reduce the effort of curating and storing millions of video frames for training.

% \begin{table*} 
% \centering
% \caption{Action detection results on ActivityNet. Note that the results of \cite{caba2015activitynet} are on the test data, ours are on the validation; The test data is not released.}
% \label{tab:actNet2}
% \begin{tabular}{c c c c c c}
% \hline \hline
% Feature  & $\alpha = 0.1$ & $\alpha = 0.2$ & $\alpha = 0.3$ & $\alpha = 0.4$ & $\alpha = 0.5$ \\  \hline 
% MF+DF+SF \cite{caba2015activitynet}  & 12.5 \% & 11.9 \% & 11.1 \% & 10.4 \% & 9.4 \%\\ 
% Ours   & \textcolor{red}{[pending]} & \textcolor{red}{[pending]} & \textcolor{red}{[pending]}  & \textcolor{red}{[pending]}  & \textcolor{red}{[pending]} \\
% \hline \hline
% \end{tabular}
% \vspace{-6pt}
% \end{table*}

%-------------------------------------------------------------------------
\section{Conclusion}

We show that utilizing web action images in training CNN models for action recognition is an effective and low-cost approach to improve performance. We also show that while videos contain a lot of useful temporal information to describe an action, and while it is more beneficial to use videos only than to use web images only, web images can provide {\em complementary} information to a finite set of videos allowing for a significant reduction in the video data required for training.

%A single web image conveys more valuable information than a single randomly selected video frame by extending content variety and weighting salient part(s) of an action in video, therefore helping the learning process of the classifier.

We observe that this approach is applicable even when different CNN architectures are used. It is also applicable using filtered image datasets or using unfiltered web crawled images. We expect that our findings should also be useful in improving the performance of the models of \cite{ng2015beyond, simonyan2014two}. 

% One interesting direction for future work is to further explore the discriminative action poses contained in web action images for better action recognition in videos, as well as for other tasks such as temporal highlighting. 

%-------------------------------------------------------------------------
{\small
\bibliographystyle{ieee}
\bibliography{egbib}
}

%-------------------------------------------------------------------------
% \begin{figure*}[h]
%   \center
%   \includegraphics[width=12cm]{average_models.png}\\
%   \caption{Suplemental Material \\
% Average accuracy improvement using web action images over the three architectures for all 101 classes of UCF101.}
%  \label{fig:avg}
% \end{figure*}

\end{document}